\documentclass[sigconf]{acmart}

\usepackage{url}
\usepackage{hyperref}
\usepackage{graphicx}
\usepackage{amsmath}
\usepackage{subcaption}
\usepackage{color}

\copyrightyear{2021}
\acmYear{2021}
\setcopyright{rightsretained}
\acmConference[SIGSPATIAL '21]{29th International Conference on Advances in Geographic Information Systems}{November 2--5, 2021}{Beijing, China}
\acmBooktitle{29th International Conference on Advances in Geographic Information Systems (SIGSPATIAL '21), November 2--5, 2021, Beijing, China}
\acmDOI{10.1145/3474717.3483651}
\acmISBN{978-1-4503-8664-7/21/11}

\begin{document}

\title[Updating Street Maps using Changes Detected in Satellite Imagery]{Updating Street Maps using \\ Changes Detected in Satellite Imagery}
\author{Favyen Bastani\textsuperscript{1}, Songtao He\textsuperscript{1}, Satvat Jagwani\textsuperscript{1}, Mohammad Alizadeh\textsuperscript{1},\\Hari Balakrishnan\textsuperscript{1}, Sanjay Chawla\textsuperscript{2}, Sam Madden\textsuperscript{1}, Mohammad Amin Sadeghi\textsuperscript{2}}
\affiliation{
  \institution{\textsuperscript{1}MIT CSAIL, \textsuperscript{2}Qatar Computing Research Institute, HBKU \\
  \textsuperscript{1}\{favyen,songtao,satvat,alizadeh,hari,madden\}@csail.mit.edu, \textsuperscript{2}\{schawla,msadeghi\}@hbku.edu.qa}
  \country{\textsuperscript{1}US, \textsuperscript{2}QA}
}
\renewcommand{\shortauthors}{Favyen Bastani, Songtao He, Satvat Jagwani et al.}

\begin{abstract}
Accurately maintaining digital street maps is labor-intensive. To address this challenge, much work has studied automatically processing geospatial data sources such as GPS trajectories and satellite images to reduce the cost of maintaining digital maps. An end-to-end \emph{map update} system would first process geospatial data sources to extract insights, and second leverage those insights to update and improve the map. However, prior work largely focuses on the first step of this pipeline: these \emph{map extraction} methods infer road networks from scratch given geospatial data sources (in effect creating entirely new maps), but do not address the second step of leveraging this extracted information to update the existing digital map data. In this paper, we first explain why current map extraction techniques yield low accuracy when extended to update existing maps. We then propose a novel method that leverages the progression of satellite imagery over time to substantially improve accuracy. Our approach first compares satellite images captured at different times to identify portions of the physical road network that have visibly changed, and then updates the existing map accordingly. We show that our change-based approach reduces map update error rates four-fold.
\end{abstract}


\maketitle

\section{Introduction}

Maintaining street maps is a labor-intensive process. As a result, many techniques have been proposed to automate parts of this process by using geospatial data sources. Current map extraction techniques~\cite{alshehhi2017simultaneous,roadtracer,cheng2017automatic,costea2017,deeproadmapper,panboonyuen2017road,vakalopoulou2015} primarily rely on satellite imagery due to its global availability, while some techniques use GPS trajectories.

A key problem with current techniques is that they are designed to infer road networks from scratch --- however, given that we already have existing high quality maps that cover the vast majority of the world, these inferred road networks are not directly useful. Instead, an end-to-end map update system must process geospatial data sources to update and improve existing digital maps.

\subsection{Map Extraction Methods Perform Poorly on Map Update}

We first consider extending current map extraction techniques for updating maps. We will show that these methods perform poorly on map update, creating many false positive updates.

Suppose that the current live digital map has a set of roads $R$. We begin by applying a map extraction method to process the most recent satellite imagery (spanning the world). This method produces another set of roads $T$ detected in the satellite imagery. Simply replacing $R$ with $T$ would not be sensible for several reasons:
\begin{enumerate}
    \item For roads that appear in both $R$ and $T$, given that $R$ is largely human-curated, it captures those roads substantially more accurately than $T$.
    \item $R$ includes roads such as tunnels that cannot be detected by the map extraction method.
    \item Roads in $R$ are labeled with rich annotations such as street names, speed limits, etc. that would be lost in the replacement.
\end{enumerate}

We could instead try to combine $R$ and $T$: if for a road segment $s$, $s \notin R$ and $s \in T$, we add $s$ to the map. (We could also remove segments $s$ where $s \in R$ and $s \notin T$, but this would prune roads that are not visible in the satellite image such as tunnels and roads occluded by buildings or trees, so we do not consider it further.)

\begin{figure}[t]
\begin{center}
	\includegraphics[width=\linewidth]{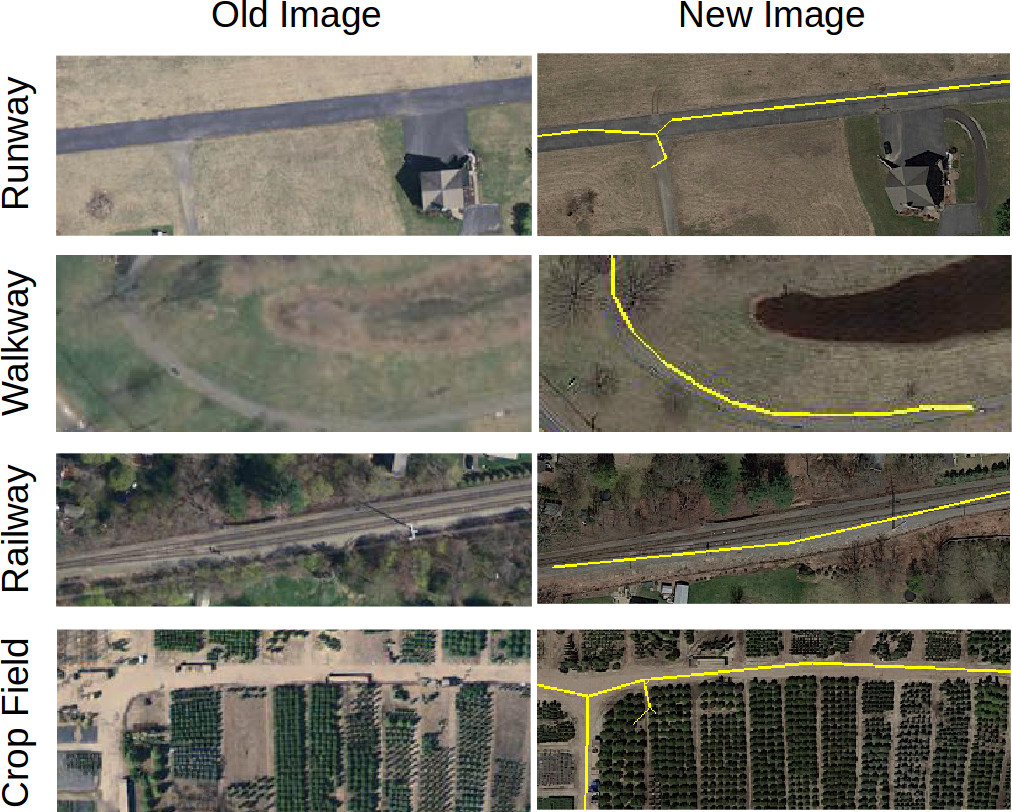}
\end{center}
	\caption{False positive detections when using MAiD to update OpenStreetMap. We show new 2017 imagery, in which the model erroneously detects roads, on the right. We include older 2015 imagery on the left to show how change detection can reduce false positives --- all of these false positives could have been eliminated by comparing satellite images over time, and determining that these are not recent road network changes.}
\label{fig:existing_fp}
\end{figure}

However, in practice, this approach yields a large number of false positives, where the map extraction method erroneously outputs many road segments in places where there are no roads.
We demonstrate this issue by using a state-of-the-art inference method, MAiD~\cite{maid}, to update OpenStreetMap~\cite{openstreetmap}.
We select a region of Massachusetts where OpenStreetMap has high coverage, and manually remove 204 groups of roads from the map that correspond to new construction between 2015 and 2017. We apply MAiD to recover these removed roads, and score its performance in terms of precision and recall comparing recovered groups of roads to the manually removed groups. At 80\% recall, MAiD yields only 67\% precision --- far too low for full automation to be a realistic option. We show examples of incorrect detections in Figure \ref{fig:existing_fp}.

Many of these detections, such as the Runway, Walkway, and Crop Field examples in Figure \ref{fig:existing_fp}, arise due to paths that are virtually indistinguishable from roads: it is restrictions set by policy governing the use of those paths, and not physical characteristics of the paths, that make them unsuitable for traversal by motor vehicles. Thus, simply improving the machine learning techniques and models used in map extraction methods is unlikely to improve accuracy; instead, a fundamental shift in approach is required.

\subsection{Key Tasks in Maintaining Maps}

To determine how accuracy can be improved, we first take a step back and identify the major challenges associated with maintaining maps. In particular, digital maps have near-complete coverage in most parts of the world: a 2015 study\footnote{See \url{https://blog.mapbox.com/how-complete-is-openstreetmap-7c369787af6e}.} found that, in 154 of 233 considered countries and territories, the length of roads in OpenStreetMap exceeded the estimated road network length reported in the CIA World Factbook, implying that the map provided excellent coverage in those countries. Then, given that digital maps have good coverage, identifying pre-existing roads constructed several years ago is not a key issue: in almost all cases, such roads already appear in the map.

Instead, \emph{the key challenge in maintaining maps is keeping the map up-to-date with changes in the physical road network}. Indeed, in the US alone, an estimated 30K km of roads are constructed each year\footnote{``Public Road Mielage'', FHWA, \url{https://www.fhwa.dot.gov/policyinformation/statistics/2013/vmt422c.cfm}.}, and map vendors spend hundreds of millions of dollars annually to keep maps up-to-date.

This presents the question: Can we develop techniques that directly tackle the key challenge of identifying physical road network changes, to substantially improve accuracy at maintaining maps over map extraction methods?

\subsection{Map Update through Change Detection}

To substantially improve accuracy, we propose leveraging a source of data largely overlooked in prior work: the progression of satellite imagery over time. By comparing satellite images captured at different times, we can hone in on portions of the road network that visibly changed over the satellite image time series. Focusing on segments that visibly changed over time enables us to disambiguate false positive road segments from genuinely new constructions: for example, all of the false positive detections in Figure \ref{fig:existing_fp} arose from curvilinear features such as walkways and crop field paths that did not undergo any recent change; thus, by processing the satellite image time series, we can determine that these are pre-existing features, and should not be added to the map.

To implement the proposed solution, we must detect changed road segments across satellite images. However, most existing change detection methods are fully supervised. They rely on collecting annotated pairs of images where change has occurred. Since newly constructed roads are rare relative to the size of the map, collecting positive examples of new roads for such a dataset is tedious and costly. Additionally, the diversity in visual appearance of roads makes change detection especially challenging. Furthermore, in Section \ref{sec:eval}, we show that prior work in unsupervised change detection exhibits low accuracy when applied for detecting new construction.

Instead, we develop a two-stage approach that requires no hand-labeling for comparing satellite images over time to detect new roads. In the first stage, we apply a novel change-seeking iterative tracing procedure to detect recently constructed roads that are missing from the existing map. Our method uses ground-truth road labels derived from the existing map dataset to avoid needing new annotations, and detects new roads that appear in an up-to-date satellite image but are not visible in an old image.

Though the first stage is effective at detecting new construction, it nevertheless yields false positive detections when occlusion and other factors yield visual differences between the old and up-to-date images despite no actual change. Thus, in the second stage, we propose a novel self-supervised change detection approach to further improve precision. We train a CNN to classify whether windows of two aligned satellite images captured at different times are cropped at the same window (matching) or at different windows (mismatched). The CNN learns to match features like road markers to determine whether two crops are matching or mismatched. To apply the model for inference, we provide it with matching crop pairs around roads detected through tracing, and we only retain detections that fool the model into classifying the pair as mismatched, suggesting the presence of new construction.

We evaluate our approach on a large-scale dataset consisting of 4800 km$^2$ of satellite imagery. We apply approaches to improve an existing map dataset, OpenStreetMap, by adding newly constructed roads to the map. At 50\% recall, our approach reduces error rates over existing state-of-the-art map inference methods four-fold, from 12\% to 3\%.
Our code and data is available at \url{https://favyen.com/mapupdate}.

In summary, our contributions are:

\begin{figure*}[t]
\begin{center}
	\includegraphics[width=\linewidth]{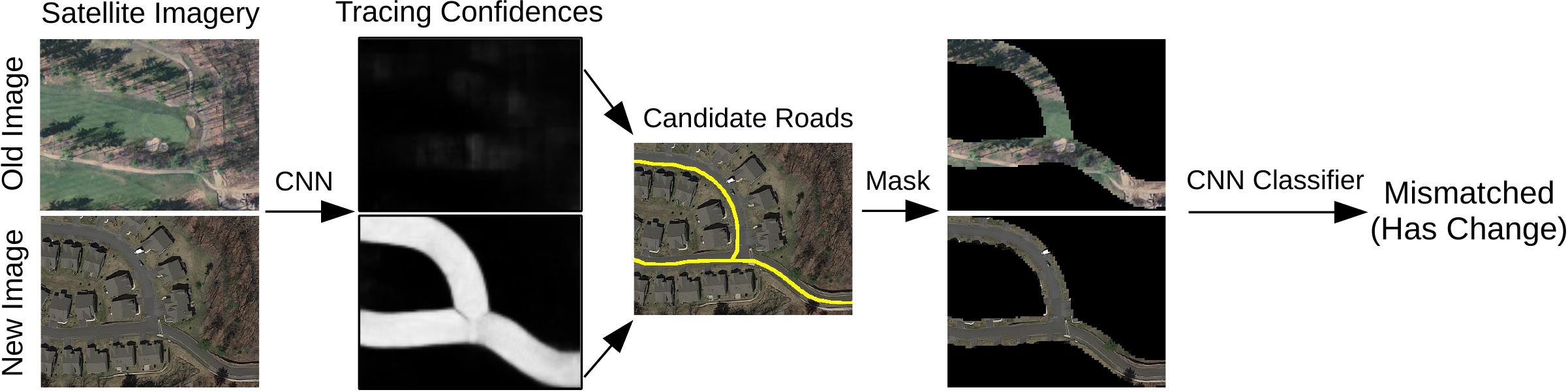}
\end{center}
	\caption{Our two-stage approach for detecting newly constructed roads in satellite imagery. Here, a pair of aligned images reflecting a new neighborhood is correctly detected.}
\label{fig:flow}
\end{figure*}

\begin{itemize}
    \item We propose a novel approach for updating street maps that leverages the progression of satellite imagery over time. In contrast to prior work that focuses on inferring all roads in a satellite image, our approach tackles the directly practical map update problem.
    \item We develop a two-stage approach for detecting new construction in satellite imagery. Our approach does not require any hand-labeling. In the second stage, we propose a novel self-supervised road-masking approach, where we train a CNN classifier in a self-supervised manner to detect change.
    \item We evaluate our approach on a large-scale dataset consisting of 3000 km$^2$ of satellite imagery in the Boston area for training and 1800 km$^2$ in northeastern Massachusetts for testing. We apply approaches to improve an existing map dataset, OpenStreetMap, by incorporating newly constructed roads into the dataset. At 50\% recall, our approach reduces error rates over existing state-of-the-art map inference methods four-fold, from 12\% to 3\%.
\end{itemize}

\section{Related Work} \label{sec:related}

\noindent
\textbf{Road Extraction.}
Automatically inferring roads from satellite imagery is a well-studied problem. Recent road extraction methods generally apply convolutional neural networks (CNNs) to segment imagery for roads~\cite{batra2019improved,costea2017,mosinska2019joint,panboonyuen2017road,yang2019road,zhou2018d}, and apply various methods to post-process the segmentation output and derive vector road network graphs. Cheng et al. apply binary thresholding, morphological thinning, and line following to extract a road network from segmentation probabilities~\cite{cheng2017automatic}. DeepRoadMapper~\cite{deeproadmapper} proposes several additional heuristic and learning-based refinement steps, including removing short edges and identifying potential missed roads.
Some methods propose alternatives to treating road extraction as an image segmentation problem.
Alshehhi et al. build the road network with a region adjacent graph that forms narrow elongated regions along roads~\cite{alshehhi2017simultaneous}.
RoadTracer~\cite{roadtracer} and PolyMapper~\cite{li2019topological} propose an iterative tracing framework to extract road networks: they train a CNN to output the directionality of roads at each pixel, and employ an iterative search guided by the CNN to trace the road network.
VecRoad extends the iterative tracing approach with a flexible step size and joint learning tasks~\cite{tan2020vecroad}, and Neural Turtle Graphics extends it with a sequential generative model~\cite{chu2019neural}.
Another recent technique, Sat2Graph, proposes a one-shot road extraction process where a CNN directly predicts the positions of road network vertices and edges~\cite{sat2graph}.

However, broadly, these approaches are unable to reason about false positive detections made by the CNN such as those in Figure \ref{fig:existing_fp}, especially for paths that appear visually similar to roads in the satellite image but are not suitable for traversal by motor vehicles. As a result, when road extraction methods are applied for the practical task of updating existing maps, they incorporate many non-road paths into the map, thereby substantially deteriorating the quality of the map data. In contrast, in most of the world where existing maps have good coverage, our method accurately keeps maps up to date with new construction by comparing satellite imagery over time, and only updating the map in areas where change is detected across images.

\smallskip
\noindent
\textbf{GPS Trajectories for Updating Maps.}
Inferring roads from GPS trajectory data has also been studied~\cite{ahmed2015comparison,biagioni,cao2009gps,davies2006scalable,prabowo2019coltrane,roadrunner,kharita}. Two works in this space, CrowdAtlas~\cite{wang2013crowdatlas} and COBWEB~\cite{shan2015cobweb}, propose \emph{map update} methods to incorporate new roads into existing maps. However, since these methods do not consider GPS time series data (comparing older trajectories to recent trajectories), they exhibit false positive errors similar to satellite image road extraction methods due to GPS noise. Additionally, due to the lack of a ground truth test set, prior work have not incorporated a quantitative evaluation of the map update portion of those methods, and instead qualitatively show results at detecting a small number of roads missing from an existing map. While it may be possible to accurately update existing maps by comparing old and recent GPS trajectory data, this has not been studied in prior work; in our approach, we focus on using satellite image time series data, since satellite imagery is globally available.

\smallskip
\noindent
\textbf{Change Detection.}
Change detection in satellite imagery has previously been studied for detecting damage from natural disasters and armed conflict. Gueguen et al. employ a semi-supervised learning approach to identify damaged regions by comparing images before and after a calamity\cite{gueguen}. However, adapting supervised and semi-supervised change detection methods for maintaining maps is difficult: annotating examples of new roads is highly time-consuming because the density of new construction is low. Unsupervised change detection methods have also been proposed~\cite{unstruct}, but we will show in Section \ref{sec:eval} that these methods exhibit poor accuracy when used to identify newly constructed roads.

\begin{figure*}[t]
\begin{center}
	\includegraphics[width=\linewidth]{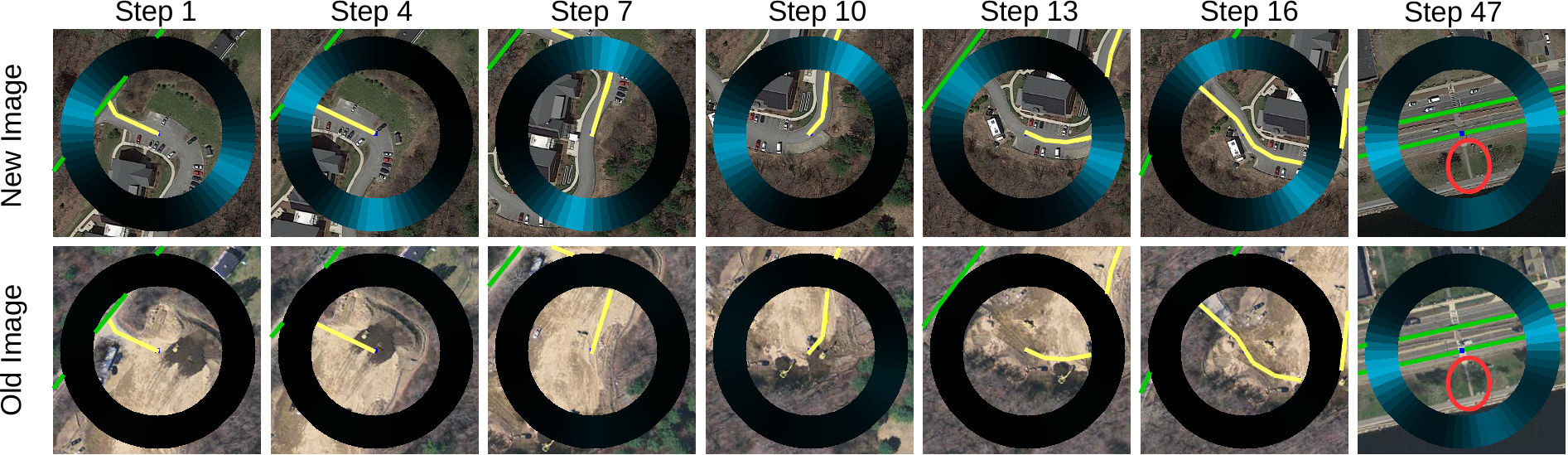}
\end{center}
	\caption{Illustration of change-seeking iterative tracing. The blue circle visualizes tracing confidences at the center pixel, with lighter colors indicating higher confidence. The existing map is green, and traced roads are yellow. On each step, we compare confidences in $M_\text{new}$ (top) against confidences in $M_\text{old}$ (bottom). On the right, this comparison helps to avoid tracing a pre-existing pedestrian path (red circle) when the model outputs similar confidences in both images.}
\label{fig:tracing}
\end{figure*}

\section{Detecting Street Network Changes} \label{sec:method}

Existing state-of-the-art map inference methods are designed to infer maps from scratch rather than update existing maps.
When applied for updating maps with new roads, these methods show poor accuracy --- the number of false positives overwhelms the few cases of new construction.
Figure \ref{fig:existing_fp} shows examples of false positive detections that arise when we apply these methods to add missing roads to OpenStreetMap.
Oftentimes, these false positives arise due to pre-existing paths such as air fields and cycling paths that appear visually similar to roads (Figure \ref{fig:existing_fp}).
We propose a novel approach that, in contrast to prior work, directly detects newly constructed roads by comparing an up-to-date satellite image against an old image.
Figure \ref{fig:flow} summarizes our approach.

Let $M_{\text{old}}$ and $M_{\text{new}}$ be old and up-to-date satellite images of the same region, and let $G$ be the road network graph of that region in the existing map. Each vertex $v \in G$ is annotated with a pixel $(i, j)$ corresponding to its location in the images, and edges correspond to roads. At a high level, in our approach, we detect new roads by identifying roads that appear in $M_\text{new}$ but not in $M_\text{old}$ or $G$. In most of the world where existing maps have good coverage, roads detected in both $M_\text{new}$ and $M_\text{old}$ likely are not roads at all, but instead non-road paths (e.g., the examples in Figure \ref{fig:existing_fp}); by comparing $M_\text{new}$ and $M_\text{old}$, our approach avoids these false positives.

We first apply a novel change-seeking iterative tracing procedure that adapts MAiD~\cite{maid} to selectively trace roads in $M_\text{new}$ that appear in neither $M_\text{old}$ nor the existing map $G$, i.e., roads that were constructed after $M_\text{old}$ was captured.
Our method traces roads along segments where a CNN model has high confidence in $M_\text{new}$ and low confidence in $M_\text{old}$.
Although this procedure improves precision over prior work, we find that it produces false positives when differences in off-nadir angle and lighting or visible non-construction activity result in a sharp increase in the CNN confidence from $M_\text{old}$ to $M_\text{new}$ despite no new roads.

Thus, we propose a novel self-supervised change detection method to automatically prune these remaining false positives in the second stage of our approach. Our method selectively identifies road network changes so that visible non-construction activity does not result in a false positive. The final road detections have high precision, and can be used to improve real-world maps through automatic merging or human validation.

In Section \ref{sec:method_segment}, we describe our method to obtain an initial set of candidate roads using change-seeking iterative tracing. We then introduce our novel self-supervised road-masking approach in Section \ref{sec:mask}.

\subsection{Change-Seeking Iterative Tracing} \label{sec:method_segment}

In the first stage, we apply a MAiD~\cite{maid} model to segment the images for tracing confidences; each tracing confidence indicates the likelihood that a road passes a pixel in a particular direction (angle). In MAiD, these tracing confidences are used by an iterative tracing algorithm to draw roads along directions with high confidence. In contrast, we develop a change-seeking iterative tracing process that avoids many of the false positives in Figure \ref{fig:existing_fp} that involve pre-existing non-road paths by comparing confidences extracted from $M_{\text{new}}$ and from $M_{\text{old}}$ to draw roads only along directions with substantially higher confidence in $M_\text{new}$, which suggests the presence of a new road.
Below, we detail each of the components in our approach.

\medskip
\noindent
\textbf{Model.} We use the MAiD CNN model architecture from~\cite{maid}. Given an image $M$, the model produces a three-dimensional matrix $P$, where $P_{i,j,k}$ is the probability that a road passes the pixel at $(i, j)$ in a direction specified by $k$. $P$ includes 64 channels, and the $k$th channel indicates the likelihood that there is a road at an angle between $k \frac{2\pi}{64}$ and $(k+1) \frac{2\pi}{64}$ from a pixel. At a pixel that falls along a straight road, channels specifying opposite directions along the road would both have high confidence. $P$ is output at one-fourth the input resolution.

\medskip
\noindent
\textbf{Training.} During training, we construct an example input by first randomly deciding whether to input $M_\text{old}$ or $M_\text{new}$, and then selecting a random 2D window in the image. We create training labels using the existing road network graph $G$. The input is a $256 \times 256 \times 3$ image (where the three channels are derived from RGB satellite imagery), and the label is a $64 \times 64 \times 64$ matrix of tracing confidences. We train the model using binary cross entropy loss, averaged across pixels and the 64 channels.

\medskip
\noindent
\textbf{Tracing Procedure.} We develop a novel change-seeking iterative tracing procedure that adapts the tracing process used in prior work~\cite{roadtracer,maid} to focus tracing on newly constructed roads and avoid inferring false positives along pre-existing non-road paths.
Figure \ref{fig:tracing} illustrates the tracing process.

Iterative tracing starts from an initial pixel known to lie on the road network, and follows directions with high confidence in $P$ at the current pixel in a depth-first search (DFS) process. The inferred road network consists of the paths that were followed during the search. 

We first compute $P_\text{old}$ from $M_\text{old}$ and $P_\text{new}$ from $M_\text{new}$. We use the existing map $G$ as a base map: our tracing procedure starts from pixels in the base map, and follows directions with high confidence in $P_\text{new}$ and low confidence in $P_\text{old}$ in a depth-first search (DFS) process. The inferred road network consists of paths that were followed during the search.
Let $G'$ be the current road network state, which we extend during tracing. We initialize $G'$ by densifying $G$ so that vertices are at most 10 m apart. We use each vertex in $G'$ as a starting pixel for tracing, and append all vertices to a DFS stack.

On each tracing iteration, we consider the pixel $(i, j)$ at the top of the search stack. We identify the highest confidence direction $k$ in $P_\text{new}[i, j]$ that has an angular distance of at least $30^{\circ}$ from any existing edges in $G'$ at $(i, j)$. This prevents re-tracing roads that are already covered by $G'$. Additionally, though, we only trace from $(i, j)$ if $P_\text{new}[i, j, k] \ge T_\text{new}$ and $P_\text{old}[i, j, k] < T_\text{old}$, i.e., if the tracing confidence in the up-to-date imagery exceeds a threshold while the confidence for the same pixel and direction in the old imagery is small. Thus, our tracing procedure only follows new roads that are not reflected in $P_\text{old}$.
If we decide to trace, then we add a new vertex $v = (i+\cos \alpha, j + \sin \alpha)$ to $G'$, where $\alpha$ is the angle corresponding to $k$, and add an edge from $(i, j)$ to $v$. We then push $v$ onto the DFS stack. Otherwise, if we decide not to trace, we pop $(i, j)$ from the stack.

We terminate tracing once the DFS stack is empty. At this point, each connected component in $G' - G$ is a candidate group of roads. By comparing $P_\text{old}$ and $P_\text{new}$ during the tracing procedure, we are able to avoid tracing along pre-existing non-road paths.

\subsection{Self-Supervised Learning for Selective Change Detection} \label{sec:mask}

\begin{figure}[t]
\begin{center}
	\includegraphics[width=\linewidth]{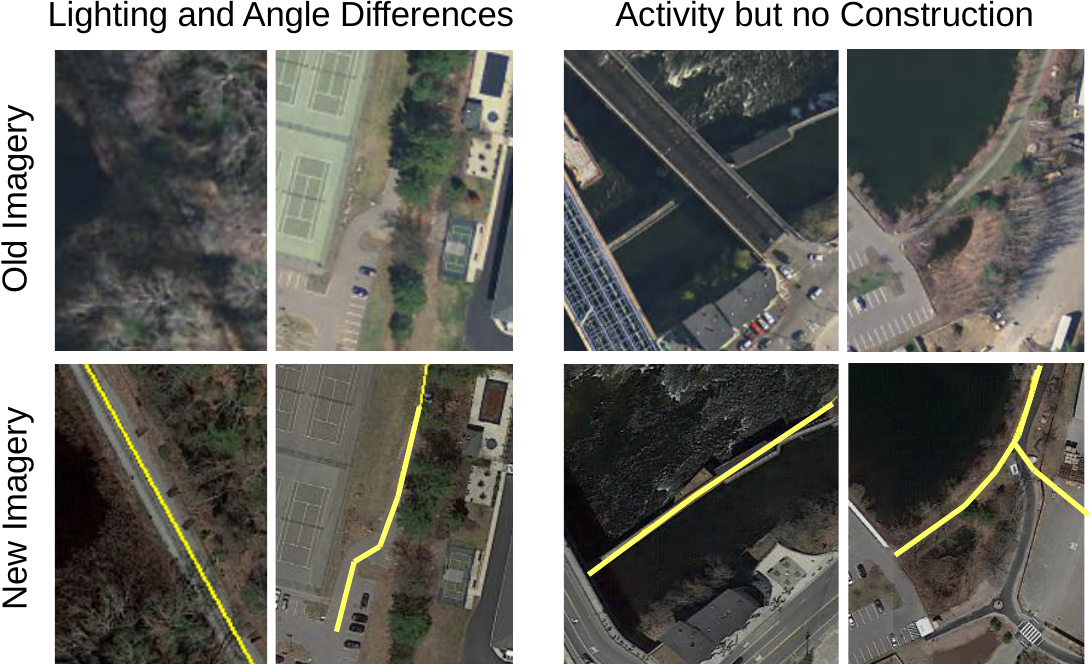}
\end{center}
	\caption{False positives after change-seeking iterative tracing. We show road detections in yellow, which correspond to a railway and three pedestrian paths. Lighting and off-nadir angle differences and visible non-construction activity lead to low tracing confidences in the old image (top) and high confidence in the new image (bottom), which results in the false positive road detections.}
\label{fig:stage1_fp}
\end{figure}

Change-seeking iterative tracing improves precision over prior work, but still produces false positives when $P_\text{new}$ reflects higher confidence than $P_\text{old}$ due to angle and lighting differences or visible activity without new construction between $M_\text{old}$ and $M_\text{new}$. Figure \ref{fig:stage1_fp} shows example false positives.
On the left, a railway and a pedestrian path are partially occluded by shadows in the old image, but are visible in the up-to-date image. On the right, although there is visible activity (bridge demolition and crosswalk painting), there are no new roads, and two walkways are incorrectly detected.

In the second stage of our approach, we apply change detection to filter the candidates generated in the first stage by pruning these false positives. However, as we will show in the evaluation, unsupervised change detection methods are unable to robustly distinguish false positives due to angle, lighting, and other aforementioned differences.
Supervised methods are also impractical: paired examples of new construction are tedious to annotate due to their low density.
Instead, we develop a novel approach that applies self-supervised learning to selectively identify road network changes.

\begin{figure}[t]
\begin{center}
	\includegraphics[width=\linewidth]{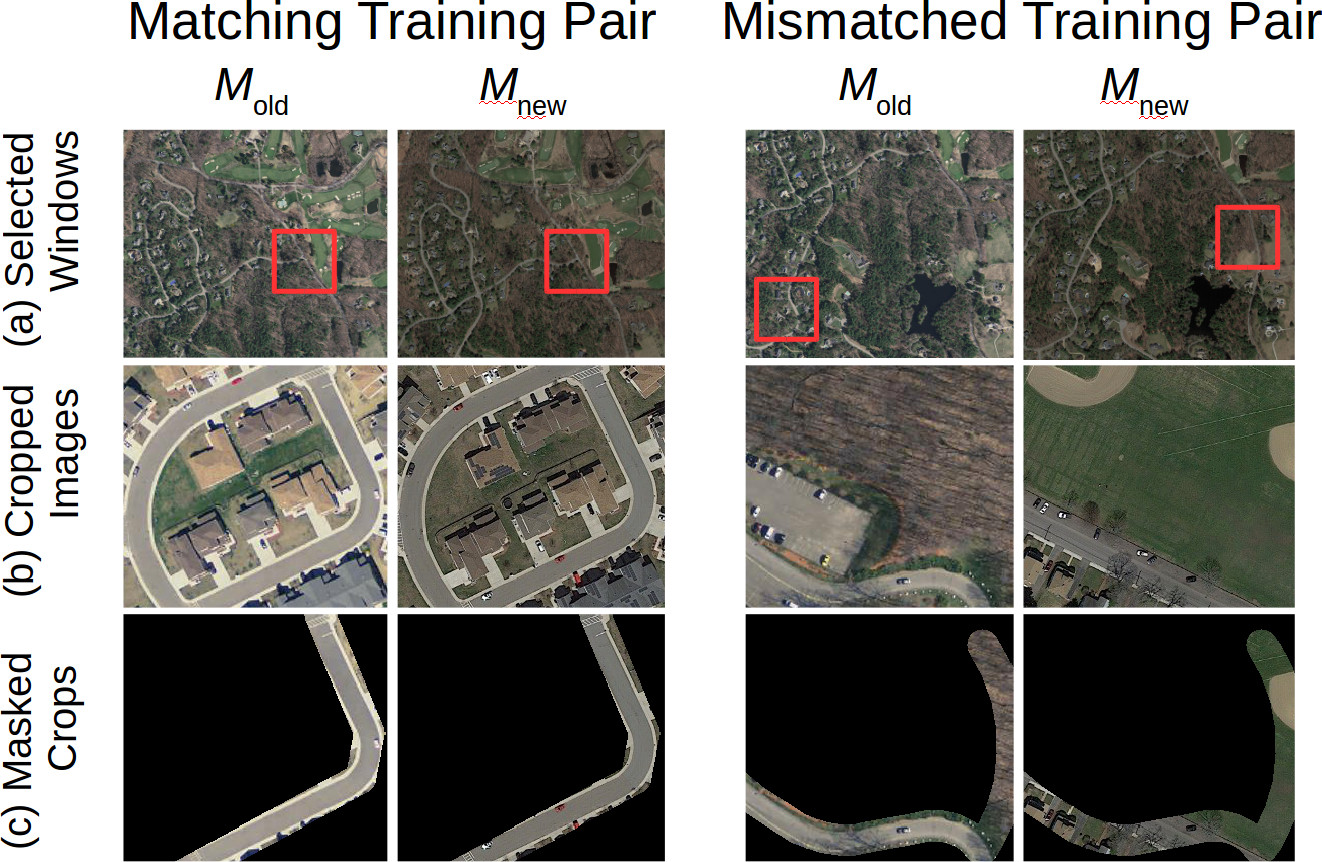}
\end{center}
	\caption{Examples of matching pairs (left) and mismatched pairs (right) that we generate during training in the second stage. (a) shows the windows selected for each example pair over a large region, (b) shows the satellite images cropped at those windows, and (c) shows the images after masking.}
\label{fig:sem_example}
\end{figure}

In our self-supervised learning procedure, we train a classifier that inputs a pair of windows of old and up-to-date images. The input may either be a matching pair, where $M_\text{old}$ and $M_\text{new}$ are cropped at the same window, or a mismatched pair, where $M_\text{old}$ and $M_\text{new}$ are cropped at disjoint windows. We train the classifier to distinguish matching pairs from mismatched pairs. We generate training examples by deciding to create a matching pair or mismatched pair with equal probability. To generate a matching pair, we randomly pick one window and crop both images at that window. To generate a mismatched pair, we randomly pick two disjoint windows. The classifier learns to match features between the images to determine whether they are taken at the same window despite differences in matching pairs such as shadows, camera angle, and non-construction activity that make unsupervised change detection methods ineffective on this task.
We show example training pairs in Figure \ref{fig:sem_example}.

To apply the filter for inference, we execute the classifier on crops of $M_\text{old}$ and $M_\text{new}$ taken at a window around connected components of roads detected during tracing. Although this pair is matching, where the crops are aligned, if there is substantial change in the images due to newly constructed roads, the crops are likely to fool the classifier into outputting a higher probability for the ``mismatched'' class. Thus, we prune candidate roads if the ``mismatched'' probability falls below a threshold.

However, in practice, new construction is often adjacent to pre-existing roads, buildings, and other structures. If the classifier observes the same structures in both the old and up-to-date windows, it can determine that the pair is matching with high confidence. Thus, we develop a masking approach that focuses the classifier on the detected candidate road.

\begin{figure}
    \centering
	\includegraphics[width=0.8\linewidth]{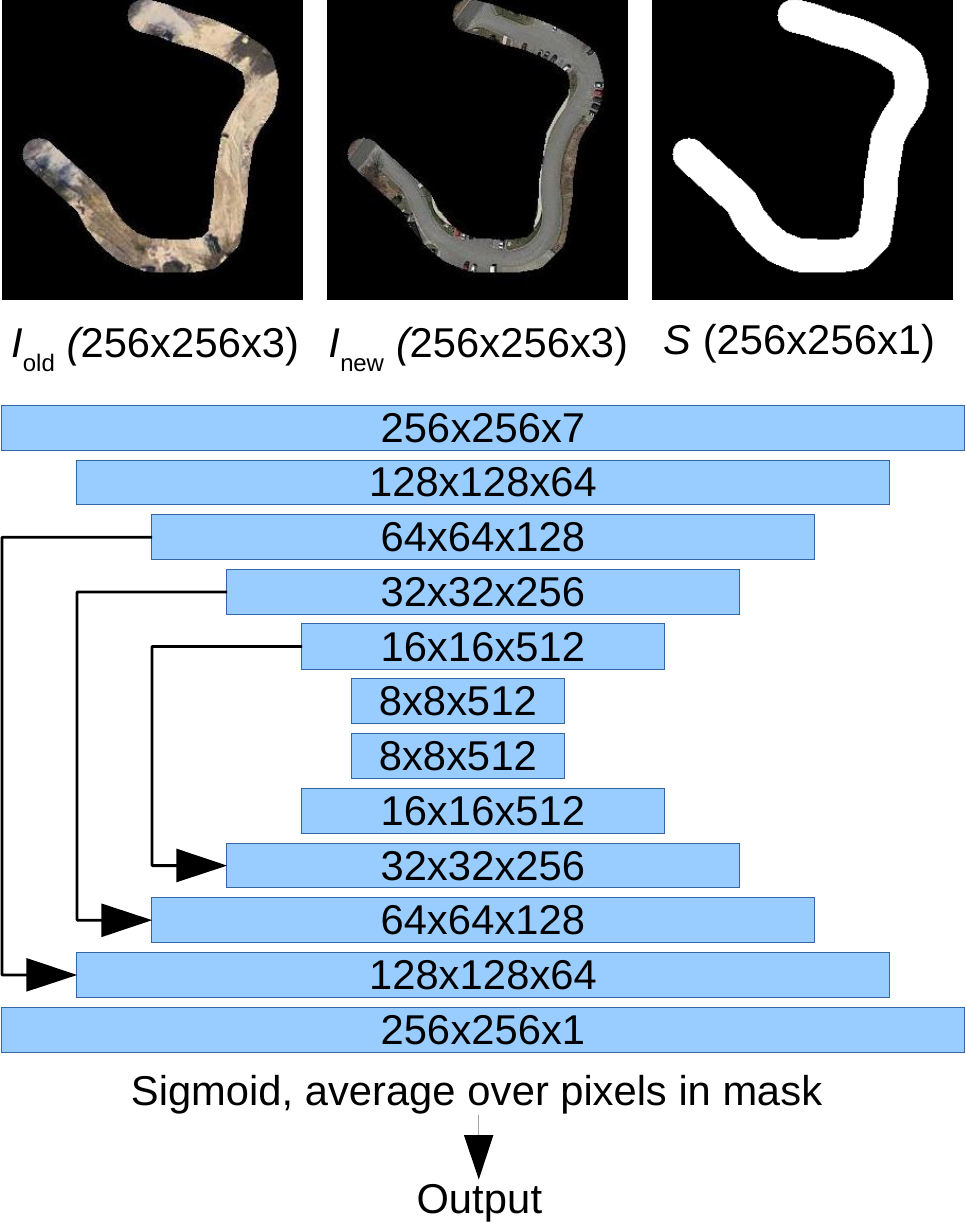}
	\caption{CNN model architecture for the classifier used in our self-supervised road-masking approach.}
    \label{fig:cnnmodel}
\end{figure}

\begin{figure}
    \centering
	\includegraphics[width=\linewidth]{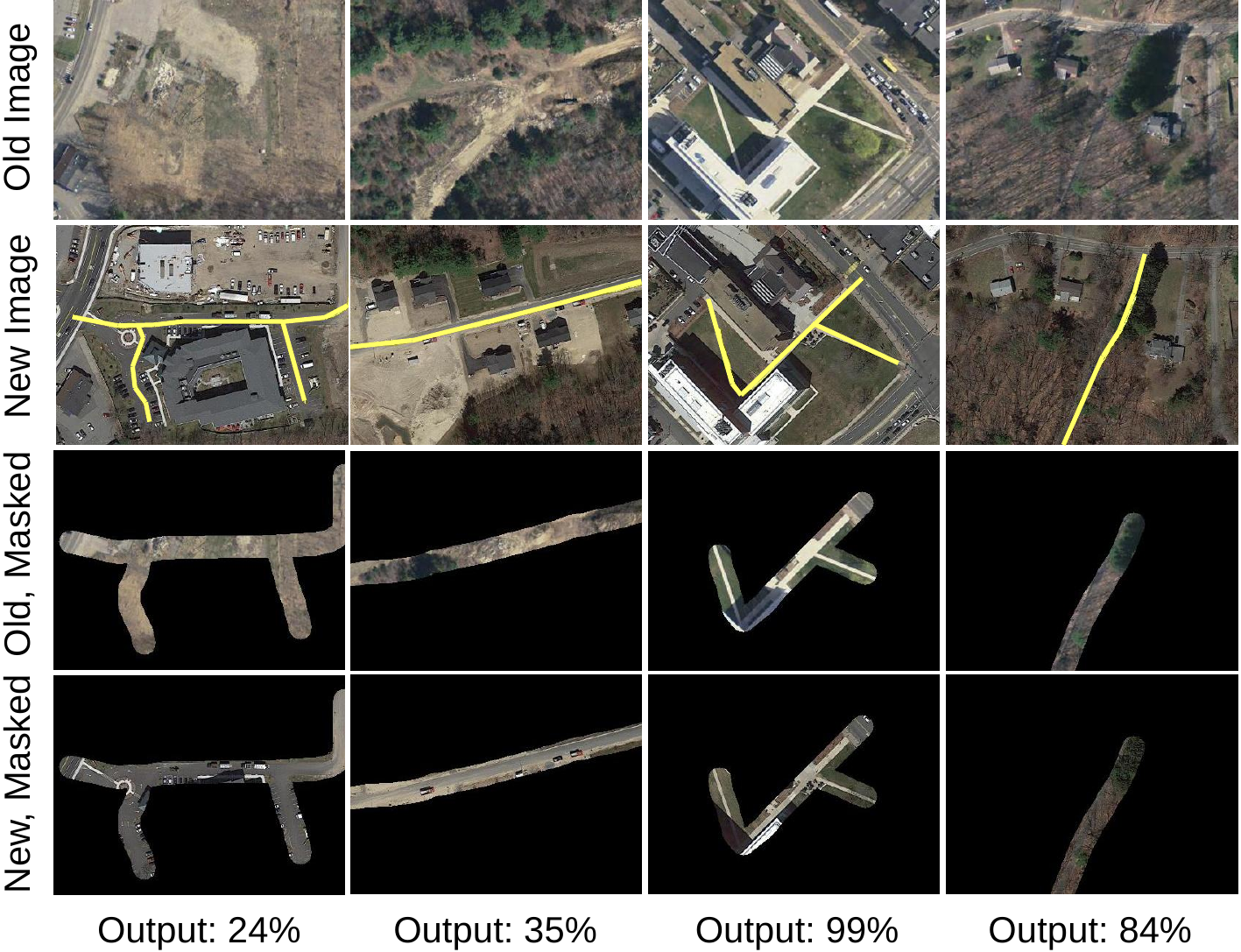}
	\caption{Classifier outputs on example groups of candidate roads inferred through change-seeking iterative tracing. At the bottom, we show the matching class probability.}
    \label{fig:sem_infer}
\end{figure}

\medskip
\noindent
\textbf{Model.} The model input consists of 7 channels: 3 channels from the crop $I_\text{old}$ of $M_\text{old}$, 3 channels from the crop $I_\text{new}$ of $M_\text{new}$, and 1 channel containing a mask $S$. $S[i,j]$ is either 0 or 1, and if $S[i,j] = 0$, then we zero the corresponding values in the crop channels, i.e., $I_\text{old}[i,j] = I_\text{new}[i,j] = 0$. Because the size of the input during inference varies based on the candidate road, we use a fully convolutional CNN architecture consisting of 6 encoder layers followed by 5 decoder layers. Figure \ref{fig:cnnmodel} shows the model architecture. The model outputs a probability at each pixel that the input example is ``matching''. We train the CNN with cross entropy loss, averaged over only pixels where $S[i,j]=1$.

\medskip
\noindent
\textbf{Training.} On each training step, we construct a matching example with 50\% probability and a mismatched example with 50\% probability. In both cases, we begin by computing the mask $S$ that we will apply to the imagery crop inputs. During inference, $S[i,j]$ will be 1 only near a group of candidate roads obtained through tracing. For effective training, $S$ must be similar to what we will provide during inference, e.g., $S$ should predominantly cover roads. Thus, we leverage $G$ to compute the mask: we randomly select a vertex $v_0$ in $G$, and perform a breadth-first-search from $v_0$ to derive a subgraph $H$ that will determine $S$. During the search, we add each traversed edge to $H$, and terminate the search once the length of the bounding box containing $H$ exceeds a threshold $T_\text{box}$. We vary $T_\text{box}$ to ensure that training examples have diverse mask sizes, since during inference, candidate groups of roads may exhibit different sizes; specifically, we pick $T_\text{box}$ uniformly between 50 m and 150 m. We set $S[i,j] = 1$ if pixel $(i, j)$ falls within 20 meters of some edge in $H$, and set $S[i,j] = 0$ otherwise.

\begin{figure*}[t]
\begin{subfigure}{.6\textwidth}
    \centering
	\includegraphics[width=0.75\linewidth]{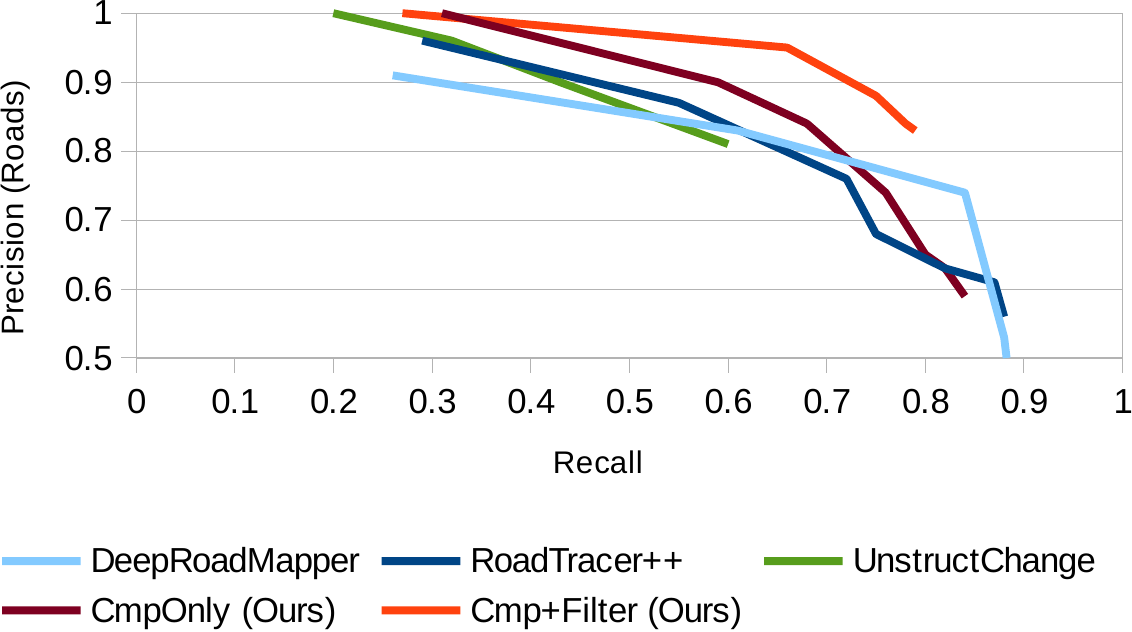}
	\caption{Precision and recall comparing detections against hand-labeled new construction.}
    \label{fig:result-road-chart}
\end{subfigure}
\begin{subfigure}{.38\textwidth}
    \centering
    \begin{tabular}{|c|r|}
        \hline
        Method & APLS \\
        \hline
        DeepRoadMapper & 0.52 \\
        MAiD & 0.49 \\
        UnstructChange & 0.48 \\
        CmpOnly & 0.51 \\
        Cmp+Filter (ours) & \textbf{0.57} \\
        \hline
    \end{tabular}
	\caption{APLS.}
    \label{fig:result-road-apls}
\end{subfigure}
\caption{Evaluation of our approach (Cmp+Filter) and the baselines on detecting newly constructed roads.}
\label{fig:result-road}
\end{figure*}

\begin{figure*}[t]
\begin{center}
	\includegraphics[width=\linewidth]{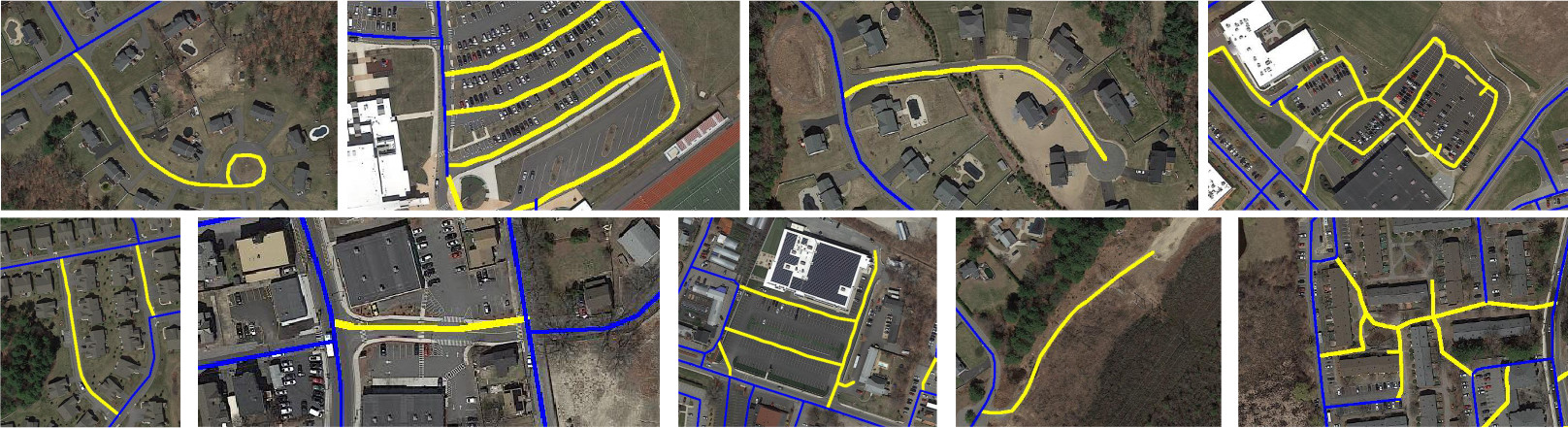}
\end{center}
	\caption{Example detections of new roads produced by our Cmp+Filter method, including two incorrect outputs on the bottom right. We show the basemap in blue and detections in yellow.}
\label{fig:result-road-example}
\end{figure*}

To create a mismatched example, we select two windows $W_\text{old}$ and $W_\text{mask}$. $W_\text{mask}$ is always a $512 \times 512$ window centered at $v_0$. We randomly pick $W_\text{old}$ so that $W_\text{old}$ and $W_\text{mask}$ are disjoint, but the distance from $W_\text{old}$ to $v_0$ is at most 2500 m. Choosing nearby but disjoint windows when creating mismatched examples is crucial as it yields more challenging examples where the old and up-to-date crops have similar style (e.g., both in suburban neighborhoods) but different semantic content. We crop $M_\text{old}$ at $W_\text{old}$, and $M_\text{new}$ and $S$ at $W_\text{mask}$.

To create a matching example, with 80\% probability, we simply crop both $M_\text{old}$ and $M_\text{new}$ at $W_\text{mask}$. However, in some cases, tracing may output roads over non-road paths; if we only train the model on matching examples where the pixels where $S[i,j] = 1$ fall on roads, the model may be ineffective on non-road inputs. Thus, with 20\% probability, we crop $M_\text{old}$ and $M_\text{new}$ at a random window $W_\text{rand}$.

Figure \ref{fig:sem_example} shows two training examples.

\medskip
\noindent
\textbf{Inference.} For each candidate group of roads $H$, we first derive a corresponding mask $S$ similar to the process during training: $S[i, j] = 1$ only if pixel $(i,j)$ falls within 20 meters of a candidate road. We crop $M_\text{old}$, $M_\text{new}$, and $S$ at a window corresponding to the bounding box of $H$ with 20-meter padding. We then compute the average probability $p$ that the model outputs over pixels where $S[i,j]=1$. Then, given a filter threshold $T_\text{filter}$, if $p < T_\text{filter}$, we prune the candidate. In Figure \ref{fig:sem_infer}, we show the average probability on several roads inferred through change-seeking iterative tracing.

\section{Evaluation} \label{sec:eval}

We evaluate our approach against existing state-of-the-art map inference and change detection methods on a task involving automatically updating OpenStreetMap with new roads.
We use 60 cm/pixel resolution satellite imagery from MassGIS from 2015 and 2017 as our old imagery $M_{\text{old}}$ and up-to-date imagery $M_{\text{new}}$, and the OpenStreetMap dataset as our road network graph. We select two disjoint sections of this dataset for training (3000 $\text{km}^2$ in the Boston metro area) and for evaluation (1800 $\text{km}^2$ in northeastern Massachusetts).

\medskip
\noindent
\textbf{Metrics.} For evaluation, we hand-annotated 204 groups of roads that appear in $M_\text{new}$ but not $M_\text{old}$. We prune these roads from OpenStreetMap to derive a road network $G$ corresponding to a map that has not yet been updated with the new imagery. We compare the methods in terms of the precision and recall on recovering the pruned roads. Each approach outputs a set of map update proposals $P = \{ H_1, \ldots, H_n \}$, where each $H_i$ is a connected component of inferred roads. The pruned roads form a set of ground truth proposals $P^{*}$. We say a proposal $H_i \in P$ matches a ground truth proposal $H_j^{*} \in P^{*}$ if the proposal bounding boxes intersect. Then, precision and recall are defined as:
$$\text{precision} = \frac{\text{\# matches}}{|P|} \quad \text{recall} = \frac{\text{\# matches}}{|P^{*}|}$$
Under this metric, each approach yields a precision-recall curve over varying confidence thresholds (e.g., in our approach, varying $T_\text{filter}$).
Because $P^{*}$ is not comprehensive, we discard a proposal $H_i \in P$ if it is a correct example of a road but does not match with any ground truth proposal.

Although the focus of our approach is on improving the precision of inferred road segments rather than improving the geometrical accuracy of those segments, we also evaluate the methods on the latter in terms of Average Path Length Similarity (APLS)~\cite{spacenet}.

\medskip
\noindent
\textbf{Baselines.} We evaluate our method against two baselines detailed in Section \ref{sec:related} that implement existing state-of-the-art road inference approaches: MAiD~\cite{maid} and DeepRoadMapper~\cite{deeproadmapper}. We apply these methods on $M_\text{new}$ to derive proposed roads, and prune proposals that correspond to roads already mapped in OpenStreetMap by pruning segments that fall within 40 m of an edge in $G$.
We also evaluate against an unsupervised satellite image change detection method, UnstructChange~\cite{unstruct}, which identifies change by comparing feature maps extracted from old and up-to-date satellite images through a VGG-19 model trained for segmentation.
To apply UnstructChange, we first obtain candidate roads through MAiD, and then eliminate candidates where the unsupervised method detects no change.

Finally, a fourth baseline, denoted CmpOnly, applies the first stage of our approach only (change-seeking iterative tracing). We denote our full approach Cmp+Filter.

\medskip
\noindent
\textbf{Results.} We show precision-recall curves on detecting new roads over varying confidence thresholds in Figure \ref{fig:result-road}, and qualitative results in Figure \ref{fig:result-road-example}.
DeepRoadMapper is unable to achieve higher than 91\% precision due to false positives, many of which correspond to pre-existing non-road paths. MAiD provides higher precision at lower recalls, but still yields only 88\% precision at 50\% recall. At 50\% recall, CmpOnly improves precision to 94\%, and Cmp+Filter further improves precision to 97\%. Thus, our method effectively prunes false positives that have lighting and angle differences or visible activity but no new roads. UnstructChange does not improve performance over MAiD: it outputs false positives due to non-construction changes (such as angle and lighting differences) between the old and up-to-date images. While our work focuses on improving the precision of road detections, Figure \ref{fig:result-road-apls} shows that our method also yields a 5\% improvement in APLS, which measures the geometrical accuracy of inferred roads.

Overall, our full approach provides near-100\% precision at reasonable recall levels. Precision is crucial because automatic integration of detections into the street map dataset is only practical if errors are rare -- otherwise, the confusion for users caused by introducing errors may outweigh the benefit from expanded map coverage.

\subsection{Updating Maps with New Buildings}

Street maps contain numerous annotations besides roads. In this section, we show that a simple adaption of our approach for detecting new construction of buildings yields high accuracy.

\begin{figure}[t]
\begin{subfigure}{\linewidth}
    \centering
	\includegraphics[width=0.8\linewidth]{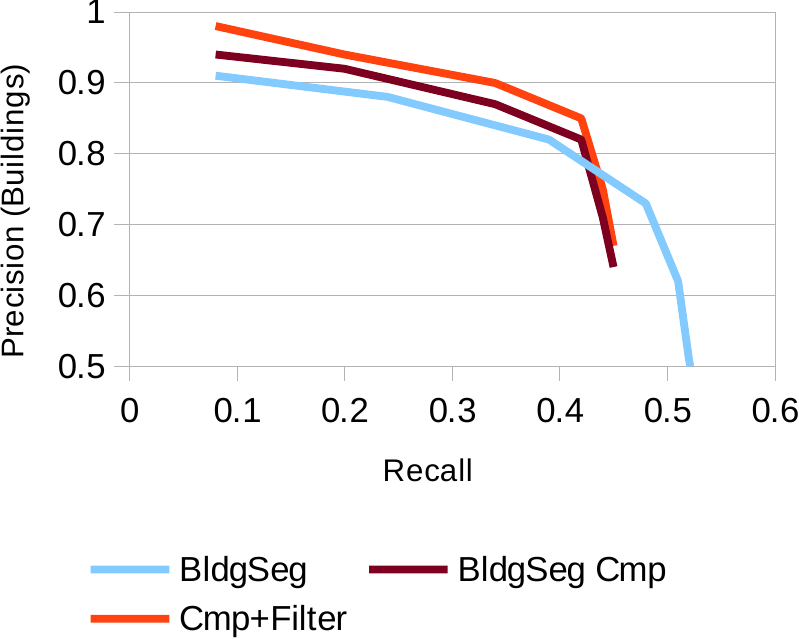}
	\caption{Precision and recall curves.}
    \label{fig:result-bldg-chart}
\end{subfigure}
\begin{subfigure}{\linewidth}
    \centering
	\includegraphics[width=\linewidth]{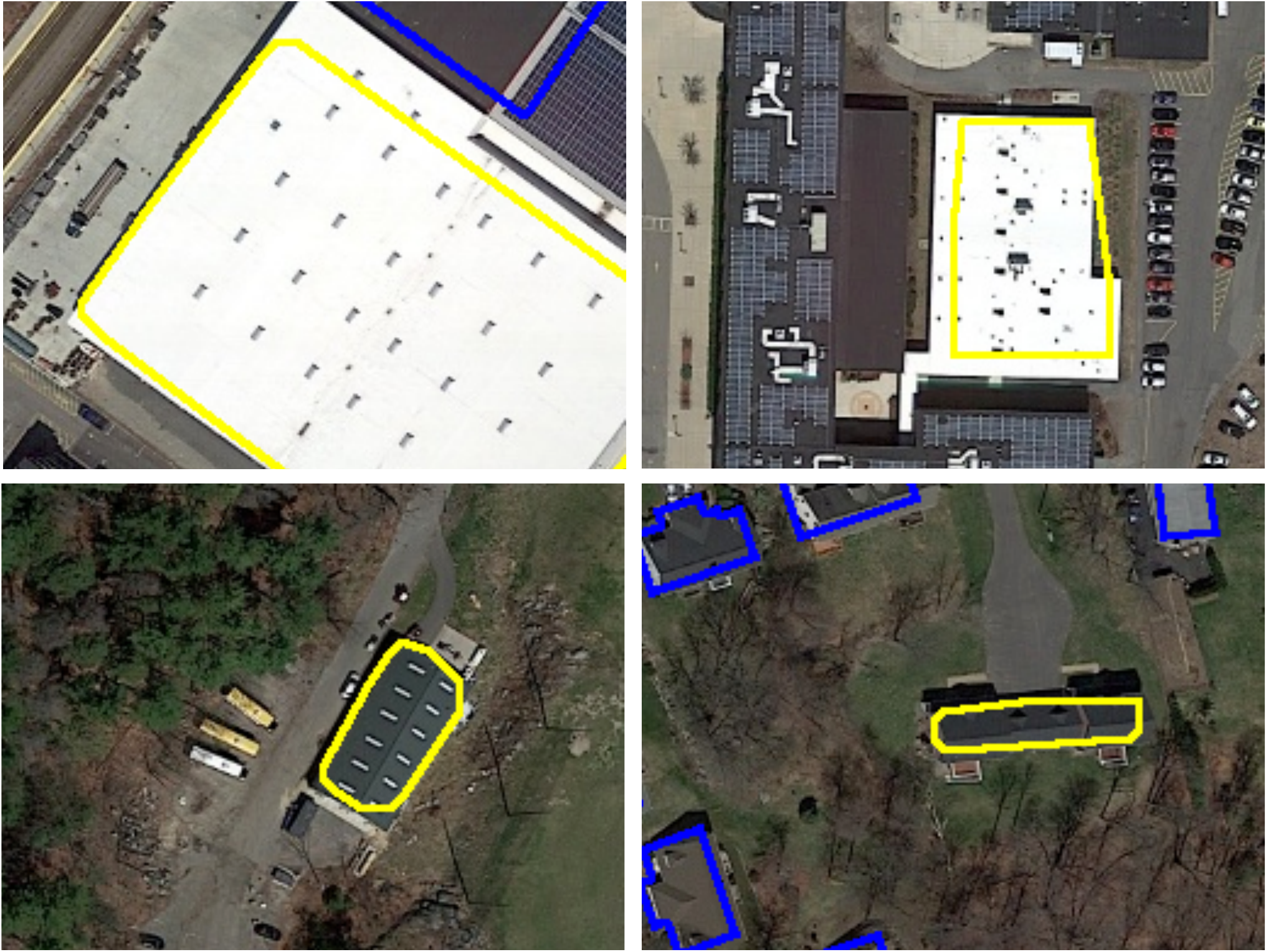}
	\caption{Example inferred buildings.}
    \label{fig:result-bldg-examples}
\end{subfigure}
\caption{Evaluation on detecting newly constructed buildings. At bottom, we show detections produced by Cmp+Filter.}
\label{fig:result-bldg}
\end{figure}

\medskip
\noindent
\textbf{Baselines.} We evaluate against two baselines. BldgSeg implements segmentation-based building extraction~\cite{building}, applying a deep CNN to segment imagery and then extracting building polygons from the segmentation probabilities. BldgSeg Cmp applies the first stage of our approach, which we adapt for buildings below.

\medskip
\noindent
\textbf{First Stage.} Our change-seeking iterative tracing method is effective at tracing road networks, but is not applicable for tracing building polygons. Instead, we apply the BldgSeg baseline to segment $M_\text{old}$ and $M_\text{new}$ for buildings, and derive candidate buildings by comparing segmentation probabilities. Specifically, we first compute segmentation probabilities $P_\text{old}[i, j]$ and $P_\text{new}[i, j]$ using the CNN at each pixel in the imagery. We then compute a binary image $B_\text{compare}$ such that $B_\text{compare} = 1$ only if $P_\text{old}[i, j] < T_\text{old}$ and $P_\text{new}[i, j] > T_\text{new}$. Finally, we extract buildings from $B_\text{compare}$. 

\medskip
\noindent
\textbf{Second Stage.} We train and apply our self-supervised model as for roads, but compute the mask $S$ by adding a fixed padding around building polygons instead of around roads.

\medskip
\noindent
\textbf{Metrics.} We evaluate the methods on precision and recall, as defined for roads. We construct a ground truth set of building polygons by removing 665 buildings from OpenStreetMap that were constructed in northeastern Massachusetts between 2015 and 2017.

\medskip
\noindent
\textbf{Results.} We show results in Figure \ref{fig:result-bldg}.
On buildings, at 30\% recall, BldgSeg yields only 85\% precision. BldgSeg Cmp improves precision to 87\%, and our approach, Cmp+Filter, yields 92\% precision. This corresponds to an almost two-fold reduction in error rate ($1 - \text{precision}$), from 15\% to 8\%.

\subsection{Removing Roads}

\begin{figure}[t]
	\includegraphics[width=\linewidth]{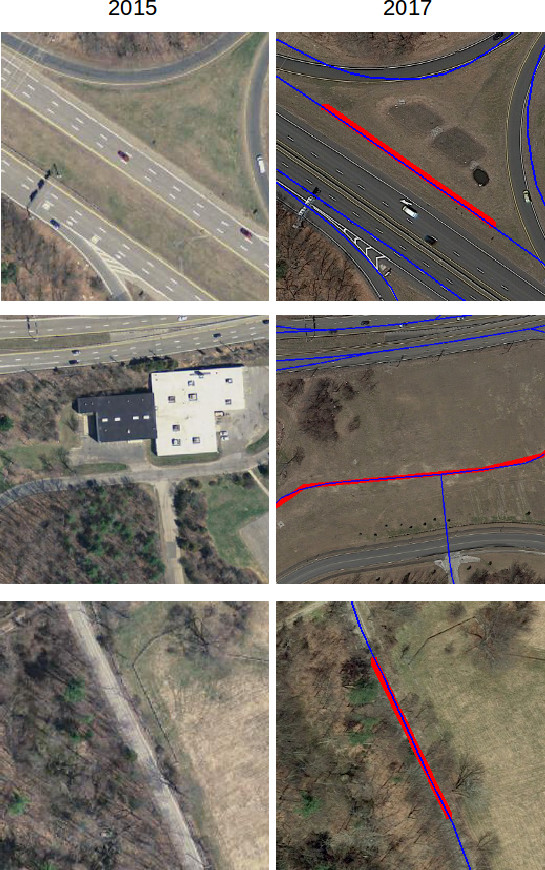}
	\caption{Applying Cmp+Filter to identify removed roads. We show old satellite imagery (left) and recent satellite imagery (right), with OpenStreetMap in blue and the detection in red. Cmp+Filter succeeds in identifying a shifted highway (top) and bulldozed road (middle), but incorrectly removes a road occluded by trees (bottom).}
	\label{fig:qual_delroad}
\end{figure}

Cmp+Filter can be applied in reverse to identify removed roads, where we identify portions of the existing map that appear in $M_\text{old}$ but not in $M_\text{new}$. Although we are unable to identify enough examples of removed roads to conduct a quantitative evaluation, we show three detections of removed roads in Figure \ref{fig:qual_delroad}. Cmp+Filter succeeds in identifying a shifted highway and bulldozed road.

\section{Conclusion} \label{sec:conclusion}

Maintaining street maps today is labor-intensive and costly. We find that existing state-of-the-art street map inference systems exhibit low precision when applied to update an existing map dataset, OpenStreetMap. By leveraging multiple satellite images collected at different times, our two-stage approach complements prior work by identifying roads and buildings that were newly constructed in the most recent image. Our evaluation on 4800 $km^2$ of satellite imagery shows that our approach is able to update existing maps to capture new construction with high precision.

\bibliographystyle{ACM-Reference-Format}
\bibliography{main}

\end{document}